\documentclass[letterpaper, 10 pt, conference]{ieeeconf}  

\IEEEoverridecommandlockouts                              
\usepackage[utf8]{inputenc}
\usepackage[T1]{fontenc}
\usepackage{dirtree}
\usepackage{booktabs}
\usepackage{bbding}
\usepackage{multirow}
\usepackage{graphicx}
\usepackage{microtype}
\usepackage{url}
\usepackage[colorlinks]{hyperref}
\usepackage{pifont}
\usepackage{blindtext}

\graphicspath{{figures/}}

\title{\LARGE \bf
MMPD: Multi-Domain Mobile Video Physiology Dataset 
}

\author{Jiankai Tang$^{1}$, Kequan Chen$^{1}$, Yuntao Wang$^{1*}$, Yuanchun Shi$^{1}$, Shwetak Patel$^{2}$, Daniel McDuff$^{2}$, Xin Liu$^{2}$\\
\small{$*$  Corresponding Author}
\thanks{$^{1}$ Jiankai Tang, Kequan Chen, Yuntao Wang, Yuanchun Shi are with the Tsinghua University, China
{\tt\small yuntaowang@tsinghua.edu.cn}.}%
\thanks{$^{2}$ Xin Liu, Shwetak Patel, Daniel McDuff are with the University of Washington, Seattle, WA, USA
{\tt\small \{xliu0, shwetak, dmcduff\}@cs.washington.edu}.}%
}

\begin{document}

\maketitle
\thispagestyle{empty}
\pagestyle{empty}

\begin{abstract}

Remote photoplethysmography (rPPG) is an attractive method for noninvasive, convenient and concomitant measurement of physiological vital signals. Public benchmark datasets have served a valuable role in the development of this technology and improvements in accuracy over recent years.
However, there remain gaps in the public datasets.
First, despite the ubiquity of cameras on mobile devices, there are few datasets recorded specifically with mobile phone cameras. 
Second, most datasets are relatively small and therefore are limited in diversity, both in appearance (e.g., skin tone), behaviors (e.g., motion) and environment (e.g., lighting conditions). In an effort to help the field advance, we present the Multi-domain Mobile Video Physiology Dataset (MMPD), comprising 11 hours of recordings from mobile phones of 33 subjects. The dataset is designed to capture videos with greater representation across skin tone, body motion, and lighting conditions. MMPD is comprehensive with eight descriptive labels and can be used in conjunction with the rPPG-toolbox \cite{liu2022deep}. The reliability  of the dataset is verified  by mainstream unsupervised methods and neural methods. The GitHub repository of our dataset: \url{https://github.com/THU-CS-PI/MMPD_rPPG_dataset}.

\end{abstract}

\section{INTRODUCTION}

Remote photoplethysmography (rPPG) is an optical technique for measuring the cardiac pulse, or photoplethysmograph (PPG), via subtle changes in light reflected from the skin \cite{mcduff2021camera}. Unobtrusive measurement of vital signs, such as heart rate, is a crucial technology for remote health monitoring and could be particularly useful for screening, and monitoring. individuals with chronic cardiovascular diseases. However, the high cost and complicated operation of traditional medical devices make regular measurements infeasible. While rPPG offers many benefits, the performance of existing video-based measurement is often brittle and can be sensitive to changes in i) appearance (e.g., skin tone), ii) the environment (e.g., lighting) and iii) activities (e.g., types of body motion). Research has shown that it is harder to extract pulse signals from individuals with darker skin tones due to the lower signal-to-noise ratio in the reflected light \cite{nowara2020meta}. Changes in lighting can significantly alter the appearance of a person's face and make it harder to detect subtle changes in reflectance due to blood flow \cite{liu2022mobilephys}. Dim or bright lighting can also lead to under or overexposure and create unwanted specular reflections, which can further obscure the signal. Motion artifacts in videos present severe challenges, and current state-of-the-art models struggle to generate precise pulse waveforms and heart rates when people are moving. Most models have not been extensively tested when users engage in naturalistic activities, such as talking or walking \cite{liu2022mobilephys,lu2021dual, mcduff2018deep}.

    \begin{table*}[t!]
        \centering
        \caption{Dataset comparison}
        \begin{tabular}{ccccccccc}
        \toprule 
        Dataset & Frames & Subjects & Camera & Sensor &  Skin Tone  & Motion & Lighting & Exercise\\
        \midrule 
        UBFC&57,420&42&Logitech C920 HD Pro&CMS50E&{\XSolidBrush}&\XSolidBrush&\Checkmark&\XSolidBrush \\
        PURE&168,120&10&eco274CVGE&CMS50E&\XSolidBrush&\Checkmark&\XSolidBrush&\XSolidBrush \\
        Scamps*&1,296,000&2800&/&/&\Checkmark&\Checkmark&\Checkmark&\XSolidBrush \\
        
        \midrule
        MMPD&1,188,000&33&Galaxy S22 Ultra&HKG-07C+&\Checkmark&\Checkmark&\Checkmark&\Checkmark\\
        \bottomrule 
        \end{tabular}\\
        \label{tab:my_label}
        \footnotesize{As different datasets contain videos with different durations, size was computed here in terms of the number of video frames. *Scamps is a synthetic dataset and therefore is not directly comparable to other datasets.} 
    \end{table*}

Public benchmark datasets are an extremely valuable resource to the scientific community; however, all datasets are finite. In the case of rPPG, existing datasets  do not contain examples that allow researchers to systematically test models across all the aforementioned dimensions (appearance, environment and activity). For example, the widely used UBFC-rPPG \cite{bobbia2019unsupervised} dataset primarily includes videos of stationary subjects with Fitzpatrick skin types 2-3. The PURE \cite{stricker2014non} dataset includes head motions that are relatively unnatural and it was also collected primarily from subjects with Fitzpatrick skin types 2-3. Finally, many of the existing public rPPG datasets were recorded using digital single-lens reflex (DSLR) cameras or devices from specialist imaging companies. This is in contrast with the most ubiquitous camera types, namely smartphone cameras.

To address gaps in existing public rPPG datasets, we introduce the multi-domain mobile video physiology dataset (MMPD). Our dataset includes 33 subjects with Fitzpatrick skin types 3-6, four different lighting conditions (LED-high, LED-low, incandescent, natural), and four different activities (stationary, head rotation, talking, and walking). All videos in MMPD are captured using mobile phones. Our paper presents the following contributions: 1) we introduce the MMPD dataset, the first public dataset that includes subjects with diverse skin types (Fitzpatrick scale of 3-6), different lighting conditions, and various real-world motion scenarios. 2) we conduct a comprehensive quantitative analysis to evaluate the performance of existing state-of-the-art neural and unsupervised signal processing methods on our dataset. Our goal is to provide researchers with a dataset that enables the development of algorithms that can handle complex and realistic scenarios, as well as address bias in camera-based physiological measurements.

    \begin{figure*}[t!]
    \centering
    \vspace{0.5cm}
    \includegraphics[width=12cm]{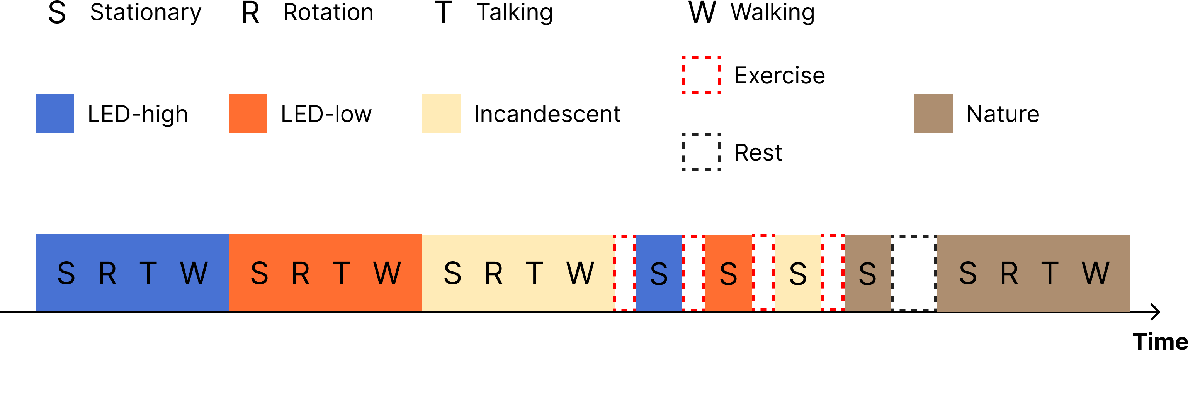}
    \vspace{-0.5cm}
    \caption{A visual illustration of our data collection protocol. Video recordings of each participant were collected under different lighting configurations (blue), activities (S, R, T, W) and before and after exercise (red vs. black box).}
    
    \end{figure*}
\section{Related Works}

There are a number of commonly used rPPG datasets (PURE~\cite{stricker2014non}, MAHNOB-HCI~\cite{soleymani2011multimodal}, BP4D~\cite{zhang2016multimodal}, VIPL-HR~\cite{niu2018vipl}, COHFACE~\cite{heusch2017reproducible}, UBFC-RPPG~\cite{bobbia2019unsupervised}, MR-NIRP~\cite{nowara2018sparsePPG}, VicarPPG-2/CleanerPPG~\cite{gudi2020real}, Scamps~\cite{mcduff2022scamps}). Some of these datasets were collected with the explicit purposes of rPPG in mind, while others were collected for generic physiological and computer vision research. From these remarkable datasets, we picked the three most frequently used datasets for analysis and comparison.

\textbf{UBFC~\cite{bobbia2019unsupervised}.} The UBFC-RPPG dataset is captured using a Logitech C920 HD Pro webcam with a resolution of 640x480 at a frame rate of 30fps. Ground truth PPG data, including the PPG waveform and heart rate, is obtained using a CMS50E transmissive pulse oximeter. The subjects are seated approximately 1 meter away from the camera with their face visible, and the experiments are conducted indoors with varying levels of sunlight and indoor lighting. While the dataset is reliable and widely used as a baseline, it has limited diversity due to the range of skin tones and motions represented.

\textbf{PURE~\cite{stricker2014non}.} The PURE database includes 60 one-minute sequences captured using an eco274CVGE camera at 30fps with a resolution of 640x480 pixels, and the PPG data is acquired in parallel by a CMS50E transmissive pulse oximeter at a sampling rate of 60Hz. The dataset is widely used due to its diversity of motions, including talking, translation, and head rotation, but it lacks variety in skin tones, real-world motion tasks, and lighting conditions.

\textbf{Scamps~\cite{mcduff2022scamps}.} The Scamps dataset provides frame-level ground-truth labels for PPG, inter-beat intervals, breathing waveforms, breathing intervals, and 10 facial actions in 2,800 video sequences. Each video is rendered using the corresponding waveforms, action unit intensities, and randomly sampled appearance properties. Although the dataset has demonstrated its potential for various applications, models trained on SCAMPS tend to have poor performance due to overfitting because of the simplistic nature of the vitals.

    \begin{table*}[t!]
    \small
	\caption{The results of unsupervised signal Processing Methods.}
	\vspace{0.3cm}
	\label{tab:evaluation}
	\centering
	\small
	\setlength\tabcolsep{3pt} 
	\begin{tabular}{r|cccc|cccc|cccc}
	\toprule
	    \textbf{Method} & \multicolumn{4}{c}{\textbf{ICA~\cite{poh2010advancements}}} & \multicolumn{4}{c}{\textbf{POS~\cite{wang2016algorithmic}  }} & \multicolumn{4}{c}{\textbf{CHROM~\cite{de2013robust} }} \\
        & MAE$\downarrow$ &  RMSE$\downarrow$ & MAPE$\downarrow$ & $\rho$ $\uparrow$ & MAE$\downarrow$& RMSE$\downarrow$ & MAPE $\downarrow$ & $\rho$ $\uparrow$ & MAE$\downarrow$& RMSE$\downarrow$ & MAPE $\downarrow$ & $\rho$ $\uparrow$ \\ \hline  \hline 
        \textbf{Skin tone} \\
        3  &8.83&12.24&12.15&0.26&\textbf{5.76}&\textbf{9.67}&\textbf{8.63}&\textbf{0.48}&6.57&10.46&9.64&0.33\\
        4 &15.16&19.81&17.60&0.12&9.06&13.51&10.37&0.23&10.57&13.80&12.69&0.15\\
        5 &14.42&17.70&20.07&-0.10&12.78&16.69&19.29&-0.03&14.65&18.91&22.29&-0.12 \\
        6 &17.14&21.52&19.77&-0.01&11.17&15.34&13.63&0.26&12.53&16.47&14.94&0.06\\
        \hline
        \textbf{Motion} \\
        Stationary &11.48&15.82&15.06&0.16&9.70&13.74&13.35&0.26&10.23&14.23&14.46&0.15\\
        Rotation   &11.75&15.88&15.35&0.06&7.50&11.99&10.85&0.40&9.28&14.02&13.10&0.16\\
        Talking    &13.14&17.18&16.50&0.20&8.05&12.60&10.87&0.30&9.31&13.51&12.26&0.25 \\
        Walking    &26.15&30.75&27.43&-0.08&17.05&21.20&18.49&-0.06&17.61&21.07&19.15&-0.12\\
        \hline
        \textbf{Light} \\
        LED-low  &12.20&16.54&15.71&0.03&9.76&14.15&13.41&0.14&10.49&14.84&14.52&0.08\\
        LED-high &11.98&15.90&15.41&0.21&7.26&11.26&10.24&0.45&9.53&13.42&13.25&0.15\\
        Incandescent &12.20&16.48&15.80&0.16&8.24&12.81&11.41&0.35&8.80&13.46&12.06&0.29 \\
        Nature &17.21&21.42&19.84&0.19&10.71&14.21&13.20&0.36&12.88&17.04&15.46&0.12\\
       \bottomrule 
       \end{tabular}
    \vspace{-0.1cm}
\end{table*}

\begin{table*}[t!]
    \small
	\vspace{0.3cm}
	\label{tab:evaluation}
	\centering
	\small
	\setlength\tabcolsep{3pt} 
	\begin{tabular}{r|cccc|cccc|cccc}
	\toprule
	    \textbf{Method} & \multicolumn{4}{c}{\textbf{GREEN~\cite{verkruysse2008remote}}} & \multicolumn{4}{c}{\textbf{LGI~\cite{pilz2018local}}  }& \multicolumn{4}{c}{\textbf{PBV~\cite{de2014improved}}  }\\
        & MAE$\downarrow$ &  RMSE$\downarrow$ & MAPE$\downarrow$ & $\rho$ $\uparrow$ & MAE$\downarrow$& RMSE$\downarrow$ & MAPE $\downarrow$ & $\rho$ $\uparrow$ & MAE$\downarrow$& RMSE$\downarrow$ & MAPE $\downarrow$ & $\rho$ $\uparrow$\\ \hline  \hline 
        \textbf{Skin tone} \\
        3 &12.37&16.48&16.67&0.15&5.99&9.83&8.10&0.45&7.94&11.36&10.96&0.38  \\
        4 &23.39&26.27&27.72&0.10&14.43&19.91&16.17&-0.17&15.87&20.50&18.34&-0.01\\
        5 &15.22&18.89&20.51&0.17&14.23&18.17&19.84&-0.02&14.62&17.77&20.17&0.08 \\
        6 &20.59&24.96&23.37&0.13&17.02&22.15&19.28&0.03&17.24&21.38&19.81&0.10\\
        \hline
        \textbf{Motion} \\
        Stationary &13.33&18.41&16.97&0.16&10.80&15.99&13.61&0.01&10.80&14.40&14.08&0.28\\
        Rotation   &16.67&20.48&21.33&0.07&9.38&14.87&12.26&0.15&11.18&16.21&14.64&0.08\\
        Talking    &17.16&21.14&21.48&0.06&11.36&16.26&13.87&0.20&13.44&17.43&16.69&0.19\\
        Walking    &29.81&34.41&31.56&0.06&25.48&30.24&26.82&0.04&25.66&30.19&27.09&0.08\\
        \hline
        \textbf{Light} \\
        LED-low &17.12&21.43&21.75&0.17&11.59&16.43&14.53&-0.01&11.91&16.09&15.16&0.11\\
        LED-high &14.71&19.13&18.76&0.08&9.74&14.91&12.46&0.17&13.01&17.22&16.95&0.09\\
        Incandescent &15.33&19.49&19.27&0.06&10.22&15.78&12.75&0.19&10.49&14.76&13.31&0.33\\
        Nature &20.07&24.74&23.19&-0.05&16.29&20.68&18.97&0.19&15.64&19.68&18.39&0.28\\
       \bottomrule 
       \end{tabular}
       \\
       \footnotesize
      MAE = Mean Absolute Error in HR estimation (Beats/Min), RMSE = Root Mean Square Error in HR estimation (Beats/Min), $\rho$ = Pearson Correlation in HR estimation.
    \vspace{-0.5cm}
\end{table*}


 \begin{table*}[t!]
    \small
	\caption{Baseline results on the MMPD datasets generated using the rPPG-toolbox \cite{liu2022deep}. For the supervised methods we show results trained on the UBFC-rPPG and PURE.}
	\vspace{0.3cm}
	\label{tab:evaluation}
	\centering
	\small
	\setlength\tabcolsep{3pt} 
	\begin{tabular}{r|cccc|cccc}
	\toprule
	    \textbf{Training Set} & \multicolumn{4}{c}{\textbf{UBFC} ~\cite{bobbia2019unsupervised}} & \multicolumn{4}{c}{\textbf{PURE} ~\cite{stricker2014non}  }\\
	    \textbf{Testing Set} & \multicolumn{4}{c}{MMPD} & \multicolumn{4}{c}{MMPD} \\
        & MAE$\downarrow$ &  RMSE$\downarrow$ & MAPE$\downarrow$ & $\rho$ $\uparrow$ & MAE$\downarrow$& RMSE$\downarrow$ & MAPE $\downarrow$ & $\rho$ $\uparrow$ \\ \hline  \hline 
        \textbf{Skin tone} \\
        3 &3.60&6.91&5.01&0.76    &\textbf{3.06}&\textbf{6.60}&\textbf{4.06}&\textbf{0.77}   \\
        4 &14.45&20.51&16.23&-0.12&8.94&15.74&9.98&0.25 \\
        5 &10.06&13.72&14.11&0.45 &12.39&16.51&16.74&0.12 \\
        6 &14.88&20.21&16.85&0.18 &15.43&20.98&17.51&0.20\\
        \hline
        \textbf{Motion} \\
        Stationary &5.34&11.17&6.32&0.56    &5.91&11.56&7.13&0.54\\
        Rotation   &11.73&16.45&15.14&0.12  &8.92&14.99&11.24&0.17\\
        Talking    &7.35&12.52&9.07&0.50    &8.32&13.71&10.21&0.42 \\
        Walking    &24.91&29.76&26.24&-0.02 &27.21&31.97&28.56&0.03\\
        \hline
        \textbf{Light} \\
        LED-low      &8.33&13.69&10.10&0.44  &7.95&13.64&9.46&0.39\\
        LED-high     &7.92&13.18&10.14&0.40	 &7.80&13.37&10.00&0.37\\
        Incandescent &8.18&13.83&10.29&0.37  &7.40&13.46&9.12&0.38 \\
        Nature       &10.41&16.77&12.37&0.36 &11.04&17.66&12.52&0.35\\
       \bottomrule 
       \end{tabular}
       \\
       \footnotesize
      MAE = Mean Absolute Error in HR estimation (Beats/Min), RMSE = Root Mean Square Error in HR estimation (Beats/Min), $\rho$ = Pearson Correlation in HR estimation.
    \vspace{-0.1cm}
\end{table*}

\section{Dataset}

In an effort to create a dataset that captures some of the diversity and complexity of videos seen in real-world applications, we recruited subjects from different countries and conducted experiments under various lighting configurations. A total of 660 one-minute videos were recorded using a Samsung Galaxy S22 Ultra, while gold-standard PPG signals were simultaneously recorded using an HKG-07C+ oximeter. In this section, we will describe the data collection protocol, data processing techniques and dataset organization.

\subsection{Data Collection} 
   \begin{figure}[t!]
    \centering
    \includegraphics[width=0.4\textwidth]{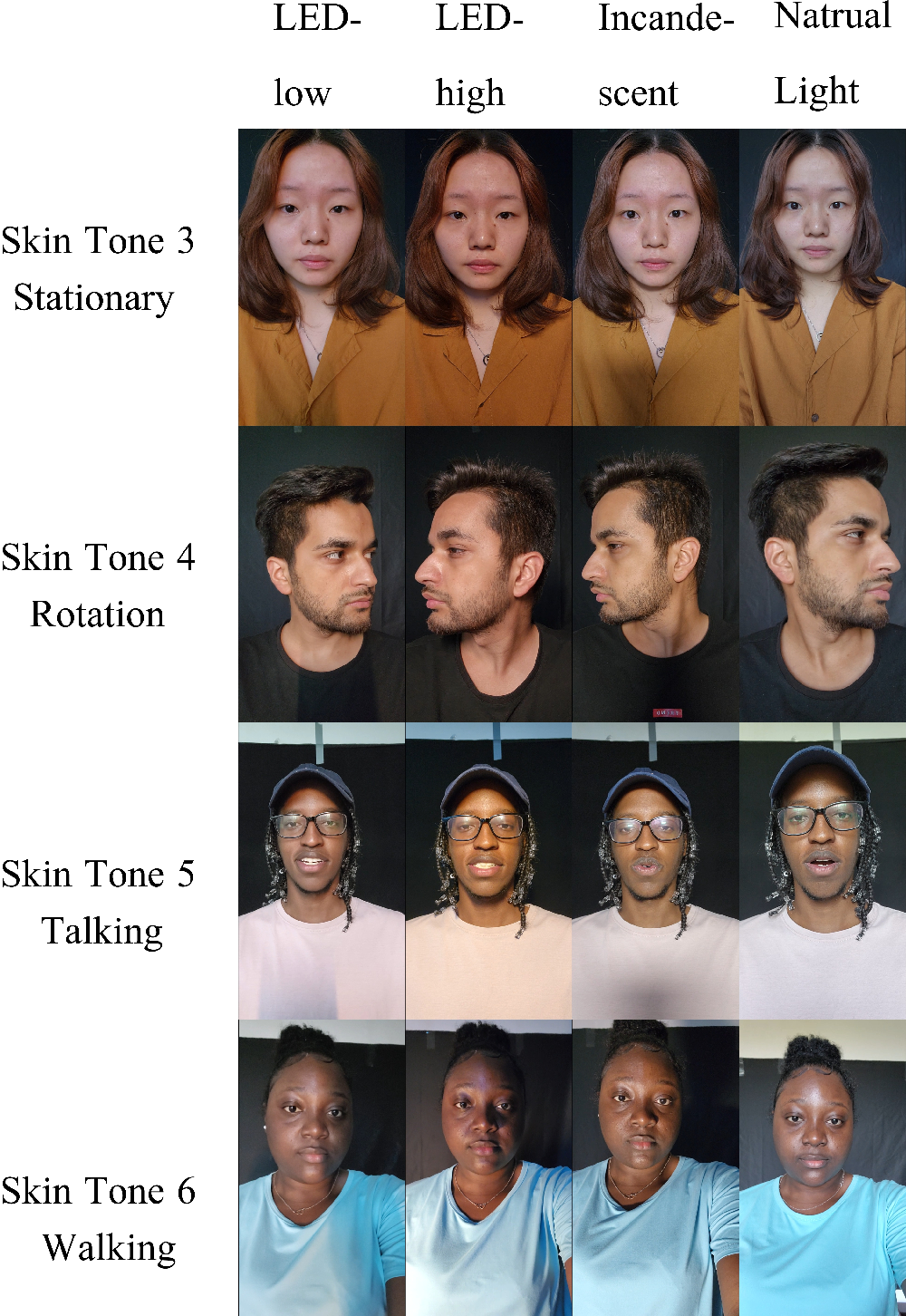}
    \caption{Sample video frames across multiple domains such as skin tones, motions and lighting conditions.
    }
    \vspace{-0.5cm}
    \end{figure}
As previously noted, lighting and motion can greatly affect the extraction of PPG signals from videos. To further study these factors, we designed an experiment that simultaneously collects face videos and finger PPG signals. The experimental procedure is illustrated in Figure 1, and all the videos were captured at a distance that allows for touch.

The experiment involved four levels of light intensity and three types of light sources, including low LED light (100 lumens on the face region), mid-level incandescent light (200 lumens on the face region), high LED light (300 lumens on the face region), and natural light (varying from 300-800 lumens intensity on the face region). For motion, we designed four tasks of varying difficulty, including remaining stationary while staring at a screen, head rotation, talking while keeping the head stationary, and taking a selfie video while holding the phone. In addition, we conducted four exercises to investigate the impact of physical activity on stationary scenarios. Subjects were asked to perform high knee lifts or other strenuous exercises to raise their post-exercise heart rate before recordings. After all exercises, subjects would take enough breaks to calm down before taking the next experiment.

\subsection{Data Processing}
To enhance the accessibility and usability of our dataset, we preprocessed the raw data and converted it into a convenient MAT file format compatible with both Matlab and Python. The videos were filmed at 30 frames per second with a resolution of 1280x720 pixels but were compressed to 320x240 pixels to facilitate storage and transmission. The PPG signals were downsampled from 200Hz to 30Hz to match the frame rate of the videos, resulting in 1800 frames per video. To enable researchers to explore the potential impact of various factors on rPPG, we assigned multiple labels, such as skin tone, gender, glasses, hair coverage, and makeup, to the dataset.

To ensure the synchronization of the videos and ground-truth PPG waves captured by different devices, we employed a Logitech Yeti microphone as an intermediary. Prior to each experiment, we recorded a chirp audio signal on both devices and then calculated the cross-correlation between the two recorded audio signals to determine the time delay between the phone and laptop. The timestamps of the oximeter were obtained through the USB COM port, allowing us to synchronize the PPG signals and video signals using two timestamps.
    
\subsection{Data Samples}
Figure 2 illustrates some samples from MMPD dataset. It includes Fitzpatrick
skin types 3-6, four different lighting conditions (LED-high,
LED-low, incandescent, natural), and four different activities
(stationary, head rotation, talking, and walking).

\section{Result and Discussion}

\subsection{Unsupervised Signal Processing Methods} 
    
Six traditional unsupervised learning methods were evaluated on our dataset ~\cite{verkruysse2008remote,pilz2018local,wang2016algorithmic,de2013robust,poh2010advancements,de2014improved}. In the skin tone comparison, we excluded the exercise, natural light, and walking conditions to eliminate any confounding factors and concentrate on the task at hand. Similarly, the motion comparison experiments excluded the exercise and natural light conditions, while the light comparison experiments excluded the exercise and walking conditions. This approach enabled us to exclude confounding factors and better understand the unique challenges posed by each task.

As shown in Table 2, for unsupervised method comparison, the LGI~\cite{pilz2018local} method performed best for relatively simple data from type 3 skin tone, while the POS~\cite{wang2017algorithmic} method had the best average performance for all conditions and robustness. For skin tone comparison, all the methods performed well on the data of skin type 3. However, for types 4, 5, and 6, most of the results showed a mean absolute error (MAE) greater than 10, indicating poor generalizability. For motion comparison, none of the models performed well for the hardest walking motion, but each model had its strengths for stationary, rotation, and talking tasks. For light comparison, there was no significant difference between the three types of artificial light, and all models performed poorly under natural light.

\subsection{Supervised Deep Learning Methods}
In this paper, we also investigated how a state-of-the-art supervised neural network performs on MMPD and studied the influence of skin tone, motion, and light. We used a pre-trained TS-CAN ~\cite{liu2020multi} model which  was trained on the UBFC \cite{bobbia2019unsupervised} and PURE \cite{stricker2014non} datasets. We used the same exclusion criteria as the evaluation on unsupervised methods.

Table 3 shows the results of the supervised neural network across different tasks. The results indicate that the neural network does not generalize well in all scenarios, as it only performs well on data from skin type 3 and with stationary tasks. This is because the training data (PURE and UBFC) only contains subjects in skin types 2-3 and mostly stationary videos. There is no significant difference between the models trained on the UBFC\cite{bobbia2019unsupervised}  and PURE\cite{stricker2014non} under the improved training framework of rPPG-toolbox \cite{liu2022deep}.

\subsection{Discussion}
The discussion of our findings indicates that the performance of supervised and unsupervised methods varies depending on the similarity of test data to training data. In our study, we found that the generalizability of supervised methods is limited when tested on subjects with skin types 4-6 or under challenging motion and lighting conditions. Conversely, unsupervised methods exhibit better generalizability as they do not rely on training. Our study also revealed that public rPPG datasets may not adequately represent real-world challenges encountered in MMPD dataset. Specifically, public datasets tend to have limited representation of skin types beyond types 2-3, mostly stationary videos, and uniform lighting conditions, leading to limited generalizability of supervised methods.
 
To improve the quality of data collection, we suggest using raw mode in the phone camera app to capture subtle changes in the face and properly positioning the phone. Additionally, minimizing complex signal processing and properly utilizing video processing tools such as ffmpeg can improve the quality of video frames and reduce time delays between the oximeter and phone. Face alignment and frame padding should also be considered, given the larger size of faces in mobile phone videos.

Overall, our study highlights the importance of properly selecting training and testing data and carefully considering the real-world challenges and limitations of data collection to improve the generalizability and accuracy of rPPG methods.


\section{CONCLUSIONS}

In this paper, we introduce the MMPD dataset, a collection of over 11 hours of video recording using a mobile phone. The dataset features subjects of four skin tones, in four motion conditions, and four lighting conditions, providing a diverse range of data for the benchmarking rPPG methods. With eight descriptive labels, the MMPD dataset aims to address the limitations of existing datasets recorded with mobile phones, particularly for videos of darker skin types and real-world motion and lighting tasks. The MMPD dataset and our evaluation of rPPG methods provide a step forward in advancing the accuracy and generalizability of this technology, with the potential to improve healthcare and other applications.

\addtolength{\textheight}{-12cm}   





\section*{ACKNOWLEDGMENT}
This work is supported by the Natural Science Foundation of China (NSFC) under Grant No. 62132010 and No. 62002198, Young Elite Scientists Sponsorship Program by CAST under Grant No.2021QNRC001, Tsinghua University Initiative Scientific Research Program, Beijing Key Lab of Networked Multimedia, and Institute for Artificial Intelligence, Tsinghua University. The experimental procedures involving human subjects described in this paper were approved by the Institutional Review Board. 



\bibliographystyle{unsrt}
\bibliography{MMPD}

\end{document}